\pdfoutput=1

\documentclass[11pt]{article}

\usepackage{EMNLP2023}

\usepackage{times}
\usepackage{latexsym}

\usepackage[T1]{fontenc}

\usepackage[utf8]{inputenc}

\usepackage{microtype}

\usepackage{inconsolata}
\usepackage{graphicx}
\usepackage{subfig}
\usepackage{adjustbox}
\usepackage{subfloat}
\usepackage{multirow}
\usepackage{booktabs}
\usepackage{makecell}
\usepackage{hhline}
\graphicspath{{images/} }
\usepackage{array}
\usepackage{enumitem}
\title{Large Language Models are Complex Table Parsers}

\makeatletter
\def\blfootnote{\xdef\@thefnmark{}\@footnotetext}
\makeatother

\author{Bowen Zhao\textsuperscript{1}, Changkai Ji\textsuperscript{2}, Yuejie Zhang\textsuperscript{1}, Wen He\textsuperscript{3}, Yingwen Wang\textsuperscript{3}, \\ \textbf{Qing Wang\textsuperscript{3}, Rui Feng\textsuperscript{123*}, Xiaobo Zhang\textsuperscript{3*}} \\
\textsuperscript{1} School of Computer Science, Shanghai Key Laboratory of Intelligent Information \\Processing, Fudan University, Shanghai 200433,  \textsuperscript{2} Academy for Engineering and Technology, \\Fudan University, Shanghai, \textsuperscript{3} Children’s Hospital of Fudan University, \\National Children’s Medical Center, Shanghai, China\\
\texttt{\{bwzhao22,22210860023\}@m.fudan.edu.cn, \{fengrui,yjzhang,hewen\}@fudan.edu.cn,} \\
\{yingwenwong,zhangxiaobo0307,wq141269\}@163.com
}
\begin{document}
\maketitle

\begin{abstract}

With the Generative Pre-trained Transformer 3.5 (GPT-3.5) exhibiting remarkable reasoning and comprehension abilities in Natural Language Processing (NLP), most Question Answering (QA) research has primarily centered around general QA tasks based on GPT, neglecting the specific challenges posed by Complex Table QA. 
In this paper, we propose to incorporate GPT-3.5 to address such challenges, in which complex tables are reconstructed into tuples and specific prompt designs are employed for dialogues.
Specifically, we encode each cell's hierarchical structure, position information, and content as a tuple. By enhancing the prompt template with an explanatory description of the meaning of each tuple and the logical reasoning process of the task, we effectively improve the hierarchical structure awareness capability of GPT-3.5 to better parse the complex tables. Extensive experiments and results on Complex Table QA datasets, i.e., the open-domain dataset HiTAB and the aviation domain dataset AIT-QA show that our approach significantly outperforms previous work on both datasets, leading to state-of-the-art (SOTA) performance.
\blfootnote{\textsuperscript{*}Corresponding author}


\end{abstract}

\begin{figure}[ht]
    \centering
    \includegraphics[width=0.48\textwidth]{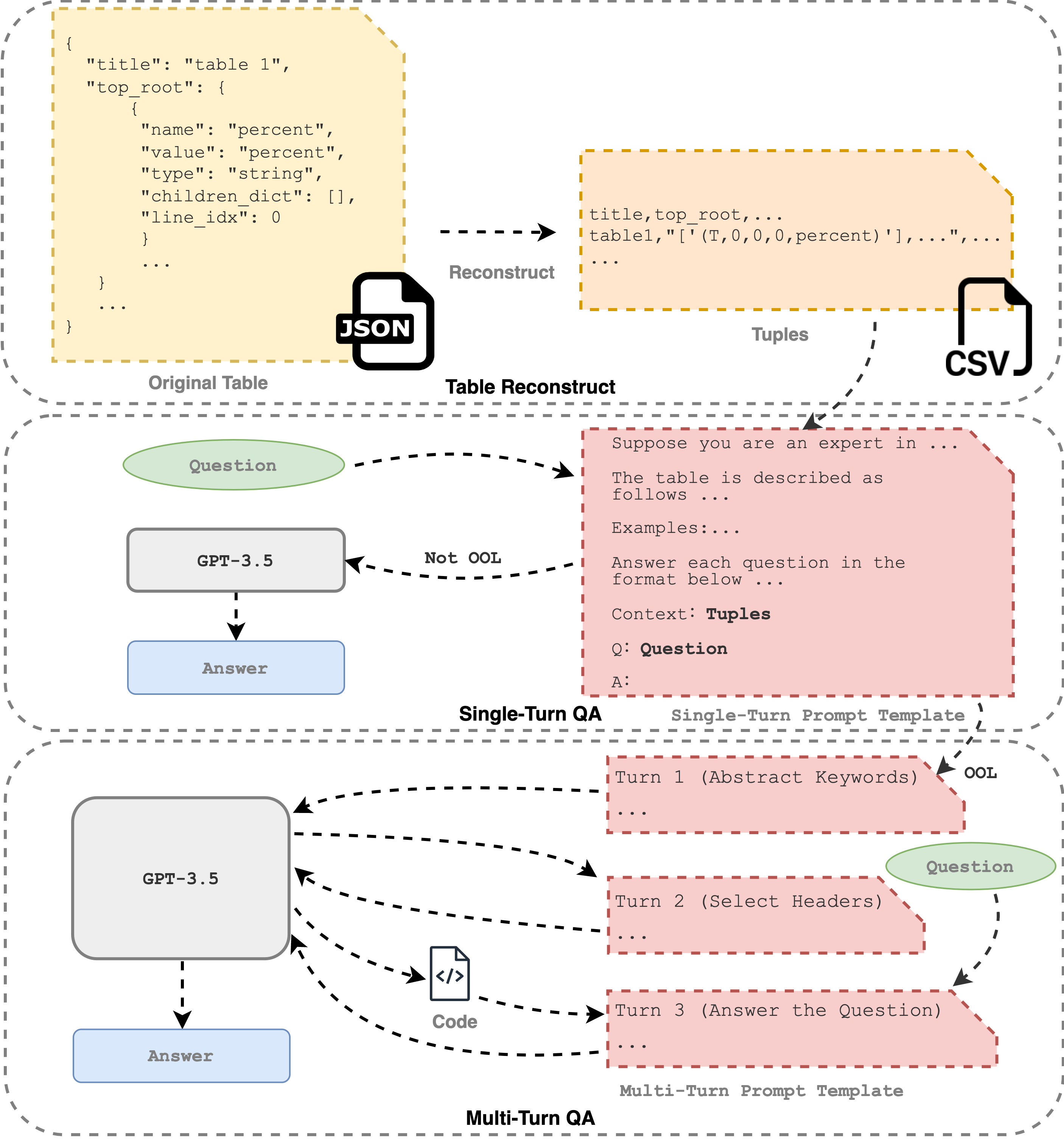}
    \caption{The overall architecture of complex table parsing. It achieves through table structure reconstruction and prompt template designing based on GPT-3.5. Texts in the pink background are prompts, where the bold text can be replaced; texts in the yellow background mean the format of the tables in the original dataset; and texts in the orange background represent the tables after refactoring. GPT-3.5 indicates the model text-davinci-003. OOL indicates that the number of tokens we requested exceeds the length limit of GPT-3.5. Code represents a piece of Python code we insert to assist multi-turn QA.}
    \label{fig:1}
\end{figure}
\section{Introduction}

Complex tables, characterized by multi-level structure headers and numerous merged cells, are a prevalent data format that includes diverse data types and intricate associations among cells \cite{complextabledef2,complextabledef1,tablesurvey1}. In this context, the Complex Table QA task emerges as an important and challenging problem within the field of NLP.

In traditional Table QA tasks, the majority of research efforts focused on simple flat tables (i.e., tables with single-level column headers and no merged cells). Earlier, a substantial number of studies focused on improving the conversion of questions into logical forms (e.g., SQLs and code) that could be directly executed on tables to retrieve answers \cite{tablesurvey1,tablesurvey2}. To this end, a wide range of strategies has been introduced, such as reinforcement learning, memory enhancement, type awareness, relationship awareness, etc. \cite{float, seq2sql, mapo, typesql, tabert}. In recent years, there has been a notable progress of Large Language Models (LLMs), such as BERT \cite{bert}, GPT \cite{gpt}, GPT-2 \cite{gpt2}, RoBERTa \cite{roberta}, GPT-3 \cite{gpt3}, and T5 \cite{t5} in the field of NLP. This enable the Table QA tasks to generate answers directly without the need for intermediate logical forms, which leverage the rich language representations and knowledge acquired through large-scale text pre-training \cite{tapas, hybridqa, mate, tableformer, llmrftr}.

While significant attainment has been made in the studies above-mentioned, their primary emphasis has been on the development of simple flat tables, overlooking the ubiquitous complex tables. Although \cite{ait, hitab} endeavored to construct QA datasets, specifically tailored for complex tables, and evaluated the performance of the SOTA Table QA models, the outcomes have not met expectations.

Most recently, with the advent of ChatGPT\footnote{\url{https://openai.com/blog/chatgpt}}, an advanced NLP model derived from GPT-3, has showcased remarkable capabilities in generation \cite{gen_chat2vis, gen2_collaboration, gen3_improving}, contextual understanding \cite{understanding1, understanding2, understanding3}, and reasoning \cite{reasoning_evaluating, reasoning2}, has profoundly impacted on the field of NLP. A multitude of research actively explores and leverages the comprehension and generation abilities of ChatGPT for QA tasks \cite{qa1, qa2_Dr, qa3_idomain, qa4_expertlevel}. However, we have not come across any relevant work that harnesses the competencies of GPT-3.5 for Complex Table QA task as yet.

In this paper we first attempt to incorporate GPT-3.5 to achieve complex table parsing, in which complex tables are reconstructed into tuples and specific prompt designs are employed for dialogues.
 As illustrated in Figure \ref{fig:1}, we reconstruct the table stored in JSON format into tuples, incorporating the hierarchical information, position information, and the value of each cell into different elements of the tuple. Subsequently, we fill in and calculate the length using the designed single-turn dialogue prompt templates. 
If the token count does not exceed the limit of GPT-3.5, we adopt a single-turn dialogue scheme for QA task. Otherwise, we utilize multi-turn dialogue scheme, in which we break down the question into sub-questions based on the logic of the single-turn prompt and introduce a code snippet to facilitate the answering process. Through careful study and meticulous experimental analysis, we conclude that GPT-3.5 can be a great parser for complex tables. 

To sum up, the contributions of this paper are:
\begin{itemize}
    \item We present a novel approach that leverages GPT-3.5 as a parser to address Complex Table QA tasks, which enhances the ability of the model to perceive the hierarchical structure of complex tables by restructuring the data format and devising prompt templates.
    \item We resolve the constraint on the input token limitation for GPT-3.5 by crafting single-turn and multi-turn dialogue prompts tailored to complex tables of different lengths, as well as incorporating code snippets for assistance in multi-turn dialogue.
    \item Extensive experiments are conducted on both pubic benchmark datasets HiTAB and AIT-QA, and the results show that our method outperforms the SOTA methods.
\end{itemize}

\begin{figure*}[t]
    \centering
    \includegraphics[width=\textwidth]{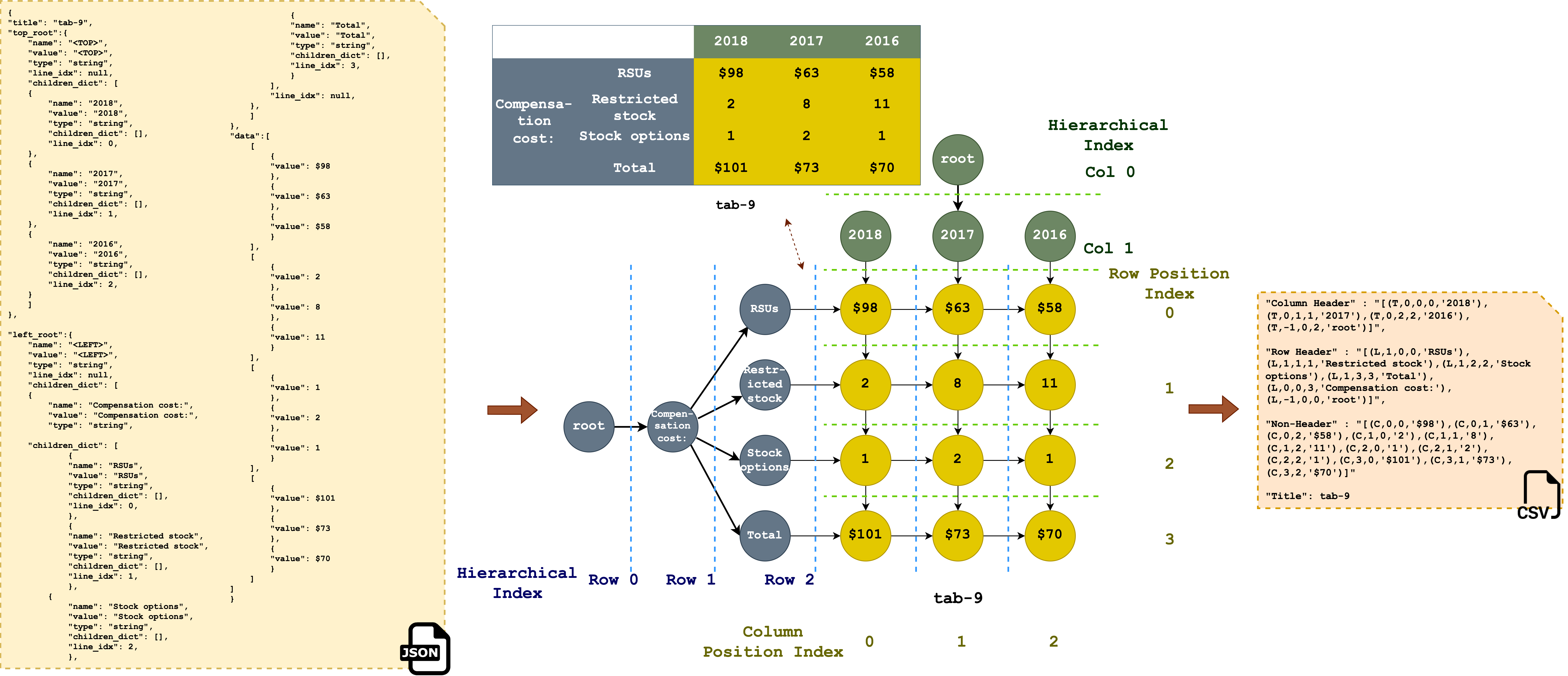}
    \caption{The illustration of table reconstruction. First, we rewrite the rows and column headers of the input JSON-formatted table into a tree structure, where, except for the root node, all nodes have a unique precursor node. Moreover, we encode the "Hierarchical-Index" of row and column headers according to the hierarchy sequence number of the tree. For example, "Col 1" indicates that 2018, 2017, 2016, the three headers are all located at the first layer of column header, and the encoding is 1. In addition, we encode the row and column position information of the header, for example, the "Column Position Index" of 2018 is 0, the "Column Position Index" of 2017 is 1, etc. Finally, we encode each cell in the table based on "Hierarchical-Index", "Row Position Index", and "Column Position Index". In the refactored table, the "Non-header" represents the cell other than the "Column Header" and "Row Header", and the "Title" represents the table title.}
    \label{fig:reformat}
\end{figure*}

\section{Related Works}
\subsection{Complex Table Question Answering}

Complex table QA tasks refer to information retrieval and answer generation on complex tables containing multi-level headers and a large number of merged cells. In previous work, most research has focused on the simple flat table dataset, such as SQA \cite{float},  Spider \cite{spider}, Hybridq \cite{hybridqa}, FeTaQA \cite{fetaqa}, etc. Recently, \cite{ait}  introduced the domain-specific Table QA dataset AIT-QA, which consists of 515 QA pairs authored by human annotators on 116 tables, and experimented with the most popular Table QA models, such as TaPAS \cite{tapas}, TaBERT \cite{tabert}, and RCI \cite{rci}, but the results were not satisfactory. \cite{hitab} also proposed a new dataset HiTab, which contains 10,672 QA pairs based on 3,597 complex tables from open-domain, and examined with Table QA models, but the results fell short of expectations.
Furthermore, they proposed a hierarchy-aware-based method, which improved the performance on complex tables with the models trained on simple flat tables.
However, the performance of these models on simple flat tables still outperformed the hierarchy-aware-based method.
Different from them, we propose to reconstruct the table format and design suitable prompt templates to fully unleash the comprehension and reasoning power of GPT-3.5, which exhibits outstanding performance on both complex table datasets.

\subsection{Prompt Engineering}
Prompt Engineering refers to making better use of the knowledge from the pre-training model by adding additional texts to the input. In the era of LLM,  \cite{cot} proposed a new method called chain-of-thought (CoT), a series of intermediate reasoning steps to unleash the power of LLMs to perform complex reasoning. Soon, \cite{zero_cot} proved LLMs to be an effective zero-shot reasoners by incorporating "Let's think step by step" before each answer. \cite{auto_cot} also proposed Auto-CoT (automatic chain-of-thought), which could generate prompts automatically. However, the power of LLMs as good reasoners has not been applied to implement complex table parsing. In this paper, we specifically focus on designing the appropriate prompts to make full use of the remarkable understanding and reasoning capabilities of GPT-3.5 to realize the structure-awareness of complex tables.

\begin{figure*}[ht!] 
  \centering
  \subfloat[An example of Single-Turn Dialogue.]{\label{fig:2(a)}\includegraphics[width=0.33\textwidth]{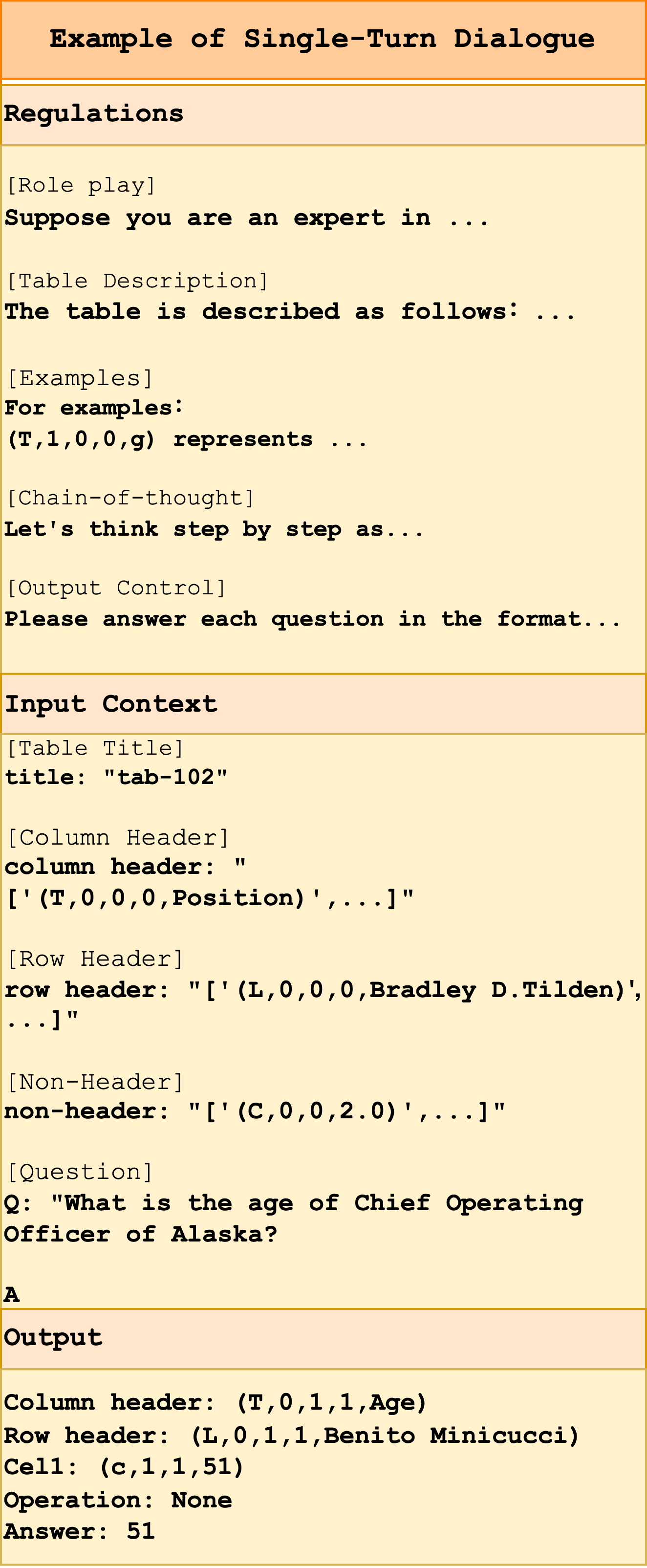}}
  \hfill
  \subfloat[An example of Multi-Turn Dialogue.]{\label{fig:2(b)}\includegraphics[width=0.63\textwidth]{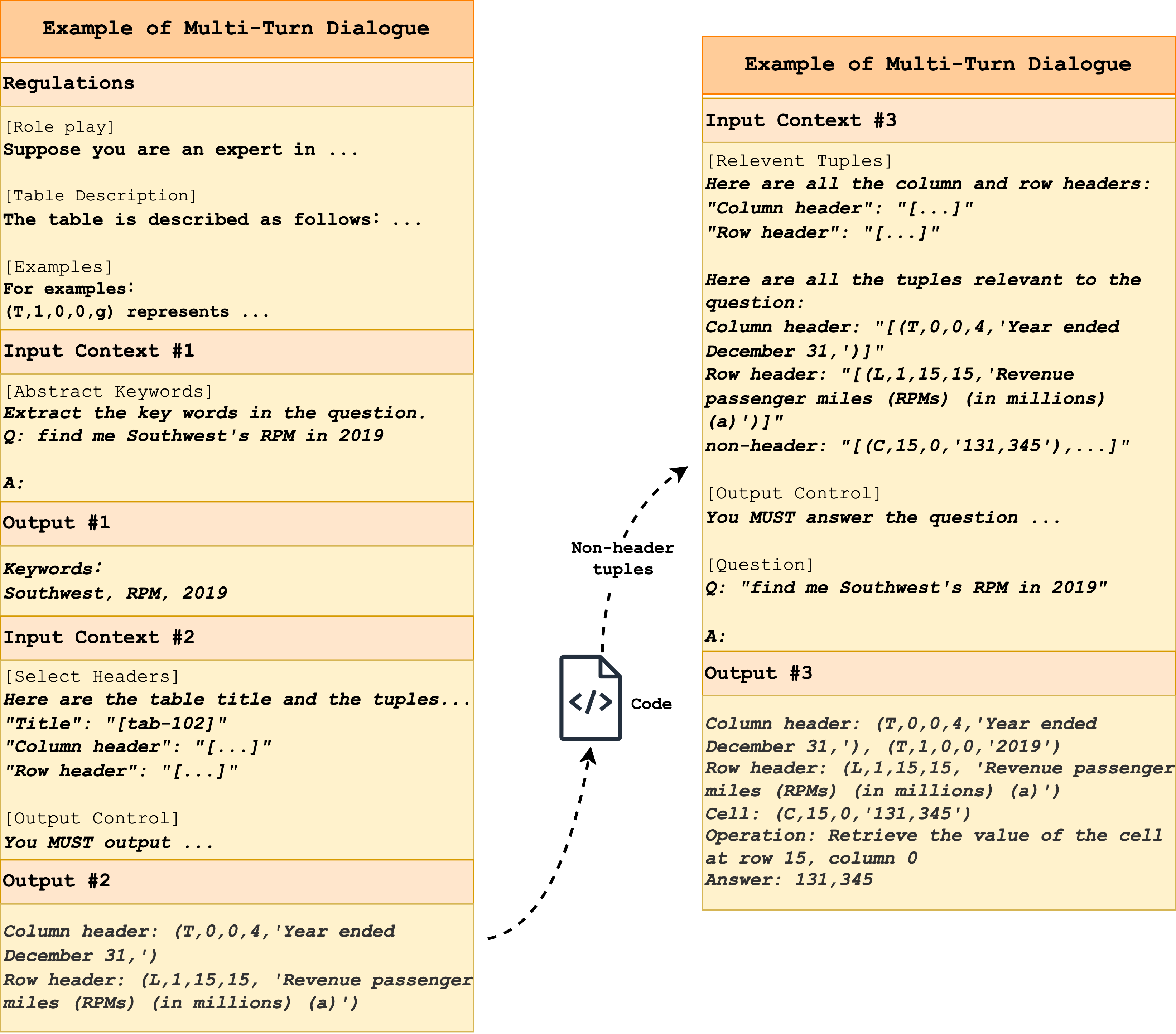}}
  
  \caption{Examples of Single-Turn and Multi-Turn Dialogue.}
  \label{fig:2}
\end{figure*}

\section{Method}
In this section, we reconstruct the table from JSON-formatted to tuples (Section \ref{Data Reformat}). To better leverage the reasoning and generation capabilities of GPT-3.5, we introduce the single-turn prompts designed for tables that do not exceed the API input limit on tokens (Section \ref{Single-Turn Prompt}) and the multi-turn prompts for tables that exceed the limit. Moreover, we add a piece of code to assist question answering in multi-turn QA (Section \ref{Multi-Turn Prompt}). Here, we use Python for coding and calling API.

\subsection{Table Reconstruction}
\label{Data Reformat}
As shown in Figure \ref{fig:reformat}, we reconstruct the table from JSON-formatted to tuples. For the row and column headers, we encode each cell as a five-tuple. The first element in the tuple serves as the label to indicate that it represents the row or column header (T stands for column header and L stands for row header). The second element denotes the hierarchical index of the cell, while the third element represents the ordinal number of the cell's starting row or column. The fourth element signifies the cell's ending row or column number, and the fifth element encapsulates the cell's value. For non-header cells, we utilize a quadruple to depict each cell. The first element in the tuple designates the label, indicating that it pertains to a non-header cell. The second element represents the row number of the cell, while the third element signifies the column number of the cell. Lastly, the fourth element encapsulates the value of the cell. For example, (L, 0, 0, 3, "Compensation cost:") means that the tuple is a row header with hierarchical-index 0, starting at row position index 0, ending at row position index 3, and the value is "Compensation cost:".

We successfully integrate the category information, position information, and value of each cell in the complex table, as well as the hierarchical information of the row and column headers into the tuple by table reconstruction.
Thus, not only the information of the whole table is clearly expressed, but also the problem of token numbers beyond the limit caused by the information redundancy of the original format can be solved.

\subsection{Single-Turn Prompt}
\label{Single-Turn Prompt}
In this section, we adopt a single-turn dialogue scheme for complex tables that do not exceed the limit of the API input token. As shown in Figure \ref{fig:2}\subref{fig:2(a)}, we reformulate the prompt of a single-turn dialogue into \textbf{Regulations} and \textbf{Input Context}. 
\begin{itemize}
    \item \textbf{Regulations} aim to dictate and guide the behavior of GPT-3.5 to make it more capable of reasoning on complex tables. In Figure \ref{fig:2}\subref{fig:2(a)}, we make the role assumption of GPT-3.5, and the reconstructed table is described in detail and illustrated with examples.
    Also, we provide a detailed description of CoT for the entire Complex Table QA task to drive GPT-3.5 starting at the top level header, to find the relevant sub-headers layer by layer and locate the non-header cells by the location information in header tuples.
    It is worth noting that in the part of "Output Control", we require the model to output the selected "Column header", "Row header", "Cell", "Operation" and "Answer" in return. In this way, we can better evaluate the reasoning and understanding abilities of the model (See Appendix \ref{Single-Turn details} for more details of single-turn prompt).

    \item \textbf{Input Context} contains a large number of fill slots, which can be filled at: \textit{"title: []", "column header: []", "row header: []", "Q: """} according to different QA pairs, where \textit{"title"} represents the title of the table, \textit{"column header"} and \textit{"row header"} refers to the column headers and row headers of the table respectively, and \textit{"Q"} denotes the question.
    Give the input in Figure \ref{fig:2}\subref{fig:2(a)} as an example, \textit{"title: [tab-}102\textit{]"} is generated by filling \textit{"title: []"} with \textit{"tab-102"}.
\end{itemize}

\subsection{Multi-Turn Prompt and Code Assistant}
\label{Multi-Turn Prompt}

In multi-turn dialogue, as shown in Figure  \ref{fig:2}\subref{fig:2(b)}, similar to the single-turn dialogue, we partition the prompt into two components: \textbf{Regulations} and \textbf{Input Context}. However, we divide the dialogue into three turns with four modules. More specifically, we not only split the \textit{Chain-of-thought} in the \textbf{Regulations} part of the single-round dialogue and assign it to the three turns of the multi-turn dialogue, but also move the \textit{Output Control} part to the last turn of the dialogue.
It is worth noting that we add a piece of code in the third module to assist the cell selection.
\begin{itemize}
    \item \textbf{In the first module} (i.e., the first prompt turn), we extract the keywords from the question.

    \item \textbf{In the second module} (i.e., the second prompt turn), we pick the relevant tuples in the row and column header tuples and record them based on the keywords we select in the first prompt turn.

    \item \textbf{In the third module} (i.e., the code assistant module), we incorporate a code snippet to facilitate the dialogue. Specifically, we pass the row and column header tuples selected in the second round of dialogue into the third module, extract the row and column position information in these tuples through the code, and retrieve the non-header tuples of the table based on this position information. All the tuples matching the location information are returned and passed to the last module. This optimization expedites the experiment by mitigating the relatively slow API calls, and significantly enhances the results by reducing the accumulation of errors resulting from the multi-turn dialogue.

    \item \textbf{In the fourth module} (i.e., the third prompt turn), we prompt GPT-3.5 for all relevant row headers, column headers, and non-header tuples, ask questions, and require the model to answer them in our prescribed format.
\end{itemize}

\begin{figure}[t]
    \centering
    \includegraphics[width=0.43\textwidth]{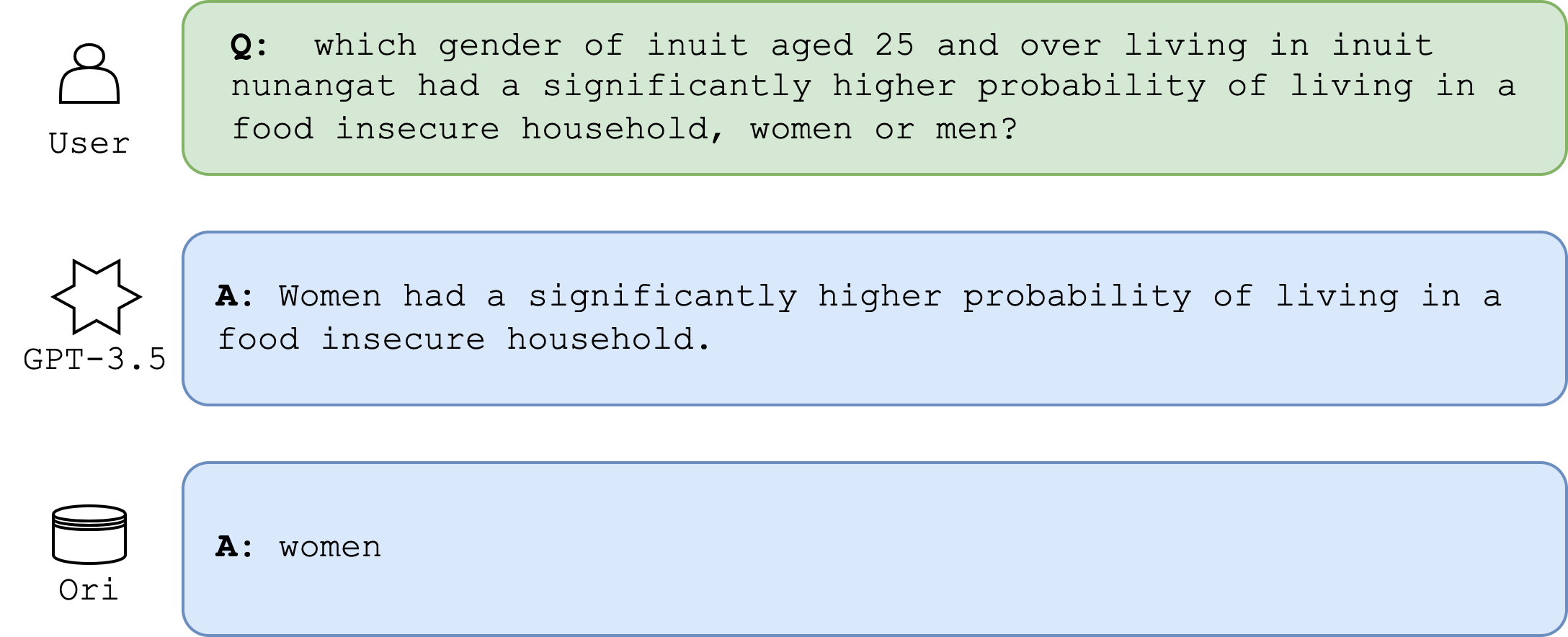}
    \caption{Schematic comparison of the answers generated by GPT-3.5 with the original answers. GPT-3.5 represents the answers generated by text-davinci-003. Ori represents the standard answer in the dataset. Texts in the green background represent the question. Texts in the blue background represent the answer.}
    \label{fig:4}
\end{figure}

\begin{table}[t]
    \resizebox{0.48\textwidth}{!}{%
    \small
    \setlength{\tabcolsep}{4pt}
    \renewcommand{\arraystretch}{1.5}
        \begin{tabular}{cccccccc}
            \hline
            \multirow{2}{*}{\textbf{Dataset}} & \multirow{2}{*}{\textbf{\makecell{Table \\number}}} & \multicolumn{4}{c}{\textbf{QA Pair}} & \multirow{2}{*}{\textbf{\makecell{Average number \\of tokens }}} & \multirow{2}{*}{\textbf{Domain}}\\
            \cmidrule(rl){3-6}
            &   & \textbf{Train} & \textbf{Dev} & \textbf{Test} & \textbf{Total} &  &   \\
            \hline
            \makecell{HiTAB} & 3,597 & 7,417 & 1,671 & 1,584 & 10,672 & 2,521 & Open domain\\
            \makecell{AIT-QA} & 116 & - & - & - & 515 & 115 & Domain specific\\
            \bottomrule
        \end{tabular}%
    }
    \caption{Dataset statistics. Average number of tokens represents the average number of tokens after tokenizing the table in the original format when calling API.}
    \label{tab:1}
\end{table}

\section{Experiment}


\subsection{Datasets}
We evaluate our approach on Complex Table QA task with HiTab \cite{hitab} and AIT-QA \cite{ait}. 
As shown in Table \ref{tab:1}, we provide the statistical results of the datasets and calculate the average length of the tokenized table. 
Moreover, it can be seen in Table \ref{tab:3} that the AIT-QA dataset is divided into four subsets. According to whether the question is related to Key Performance Indicators (KPIs), the dataset is divided into "KPI-driven" and "Table-driven". Similarly, AIT-QA is also divided into "Row header hierarchy" and "No row header hierarchy", according to whether the answer relies on the row header hierarchy.

It is worth noting that the data in HiTAB comes from statistical reports across 28 different fields and Wikipedia, offering rich table information with complex structures. In contrast, the data in AIT-QA are all selected from airline companies, and the information and hierarchical structures contained in the tables are relatively simple compared to HiTAB.

\subsection{Baselines}
Since we have not found any further work related to HiTAB and AIT-QA, we compare our work with the evaluation results in their original paper, including 1) \textit{supervision-based method:} MAPO \cite{mapo}, TaPas \cite{tapas}, MML \cite{mml} and REINFORCE \cite{reinforce}; and 2) \textit{zero-shot-based methods:} TaBERT \cite{tabert}, TaPas \cite{tapas} and RCI \cite{rci}.
\subsection{Evaluation}
Because of the presence of a substantial number of non-numerical type answers in the datasets, direct alignment or similarity evaluation cannot effectively assess our method. As illustrated in Figure \ref{fig:4}, it is evident that the answers generated by GPT-3.5 have essentially the same meaning as the original answer.
However, there exists a notable difference in their representation.
The model-generated answers tend to provide more intricate details, while the original answers exhibit a more concise nature.
Therefore, we use Accuracy as our evaluation metric following \cite{hitab}, which indicates the  percentage of QA pairs with correct answers, to align whether the generated answers are equivalent to the original answers in the case of a specific question and context (i.e., table content).

\subsection{Model}
We conduct experiments mainly with the text-davinci-003 from OpenAI, which is improved on GPT-3 and can understand as well as generate natural language or code. Text-davinci-003 supports 4,097 input tokens at most, which means that the combined count of the input tokens and the generated tokens can not exceed 4,097.

\subsection{Results}
\begin{table}[t]
\resizebox{0.48\textwidth}{!}{%
    \centering
    \begin{tabular}{llcc}
    \hline
     & \textbf{Method} & \textbf{Dev} & \textbf{Test}\\
     \hline
     Weak Supervision & MAPO \textsubscript{w. original logical form} & 31.9 & 29.2 \\
     \cite{hitab} & TaPas \textsubscript{w/o. logical form} & 39.7 & 38.9 \\
      & MML \textsubscript{w. h.a. logical form}  & 38.9 & 36.7 \\
      & REINFORCE \textsubscript{w. h.a. logical form}  & 42.7 & 38.4 \\
      & MAPO \textsubscript{w. h.a. logical form}  & 43.5 & 40.7 \\
      \hline
      Partial Supervision & TaPas \textsubscript{w/o. logical form} & 41.2 & 40.1 \\
      \cite{hitab} & MML \textsubscript{w. h.a. logical form}  & 45.4 & 45.1 \\
      & REINFORCE \textsubscript{w. h.a. logical form}  & 44.0 & 39.7 \\
      & MAPO \textsubscript{w. h.a. logical form}  & 44.8 & 44.3 \\
      \hline
      Zero-shot & Ours \textsubscript{w. GPT-3.5} & \textbf{49.0} & \textbf{50.0} \\
      \hline
    \end{tabular}
}
    \caption{Accuracy on dev/test of HiTAB. h.a. stands for hierarchy-aware.} 
    \label{tab:hitab_result}
\end{table}

\begin{table}[t]
    \resizebox{0.48\textwidth}{!}{%
    \centering
    \setlength{\tabcolsep}{4pt}
    \renewcommand{\arraystretch}{1.5}
    \begin{tabular}{lcccccc}
    \hline
    \multirow{2}{*}{\textbf{Data subset}} & \multicolumn{3}{c}{\textbf{Models}} &\multicolumn{3}{c}{\textbf{Ours}} \\
    \cmidrule(l){5-7} \cmidrule(rl){2-4}
     & \textbf{TaBERT} & \textbf{TaPaS} & \textbf{RCI} & \textbf{Single-turn} & \textbf{Multi-turn} & \textbf{All} \\
    \hline
    KPI-driven & 41.37 & 48.26 & 60.00 & \textbf{76.92} & 66.67 & 74.48  \\
    Table-driven & 31.08 & 50.0 &  48.64 & 76.67 & \textbf{80.00} & 71.84  \\
    Row header hierarchy & 21.92 & 47.26 & 45.89 & \textbf{61.72} & 61.11 & 61.64 \\
    No row header hierarchy & 38.75 & 50.39 & 54.20 & \textbf{82.23} & 78.95 & 81.84 \\
    \hline
    Overall & 33.98 & 49.32 & 51.84 & \textbf{76.73}  & 70.27 & 76.26   \\
    \hline
    \end{tabular}
    }
    \caption{Accuracy on AIT-QA. We compare with other baselines: TaBERT, TaPaS, RCI \cite{ait}.}
    \label{tab:3}
\end{table}

\begin{table}[t]
\centering
\resizebox{0.4\textwidth}{!}{
    \centering
    \begin{tabular}{lcccc}
    \hline
     & \textbf{Train}& \textbf{Dev} & \textbf{Test} & \textbf{Overall} \\
     \hline
     Single-turn &  \textbf{52.0} & \textbf{51.1} & \textbf{51.1} & \textbf{51.7} \\
     Multi-turn & 40.8 & 43.5 & 47.0 & 42.1 \\
     \hline
     Overall & 48.9 & 49.0 & 50.0 & 49.3 \\
     \hline
    \end{tabular}
}
    
    \caption{Accuracy of different dialogue schemes on train/dev/test/overall of HiTAB.}
    \label{tab:cmp}
\end{table}

\subsubsection{Main Results}

The results on HiTAB are shown in Table \ref{tab:hitab_result}.
Specifically, the absolute accuracy improvement of 5.5 on the Dev set and 9.3 on the Test set. Specifically, the absolute accuracy improvement of 5.5 on the Dev set and 9.3 on the Test set can be observed if compared to the previous best weak supervision method MAPO with the hierarchy-aware logical form on HiTAB. 
Among the partial supervision methods on HiTAB, our approach still outperforms previous works by a large margin.
Table \ref{tab:3} reports the results on AIT-QA.
In the context of zero-shot learning, the accuracy of our method outperforms TaBERT, TaPas, and RCI by 42.28, 26.94, and 24.42 on the overall dataset, respectively.
Combining the results on these two datasets, it can be demonstrated that GPT-3.5 can achieve the parsing of complex tables given appropriate data format and prompt.


\subsubsection{Results on HiTAB} 

As shown in Table \ref{tab:hitab_result}, our method outperforms all previous methods by over 3.5 in accuracy across the board.
At the same time, it can be seen from Table \ref{tab:cmp} that, the results of the single-turn dialogue are significantly higher than the overall results. These indicate that our approach of leveraging the Large Language Model as a parser for complex tables is effective.
In multi-turn dialogue, our method achieves 43.5 and 47.0 accuracy on the Dev and Test sets, respectively. 
Notice that, our work is built upon the framework of zero-shot learning, differing from the supervised methods utilized in previous studies. 
Even if our results may not consistently surpass those of these previous works, we assert our method is still comparable to the SOTA as reported in the original paper \cite{hitab}
Although the training set is also applicable, we find it performs slightly worse partially due to the dataset bias.

\subsubsection{Results on AIT-QA}
We divide the AIT-QA dataset into four subsets according to the method in the original paper \cite{ait}, and analyze them separately, as shown in Table \ref{tab:3}.
 Overall, compared to other models trained on tabular data, our method soundly outperforms all previous works. The overall accuracy of single-turn dialogue reaches 76.73, and of multi-turn dialogue, it reaches 70.27, which is a significant leap compared to the 51.84 accuracy of the the previous best method RCI \cite{rci}.
 Furthermore, our method achieves the best results on two subsets "KPI-driven" and "Table-driven", respectively, as well as on two subsets determined by whether or not the answer relies on the row header hierarchy. It is noteworthy that, on the AIT-QA dataset, GPT-3.5 still exhibits a trend where the accuracy of single-turn dialogue is generally higher than that of multi-turn dialogue. We consider that this is due to the inability of multi-turn dialogue to preserve historical information, resulting in information loss and the accumulation of errors throughout multiple interactions.

In addition, an absolute improvement over previous work is achieved on all subsets of AIT-QA. The results on subsets "Row header hierarchy" and "No row header hierarchy" show that the accuracy of answers that do not rely on the row header hierarchy is significantly higher than those that do rely on it. We attribute this to a bias in the attention of GPT-3.5 to row and column headers.

{\small
\begin{table}[]
    \tiny
    \resizebox{0.48\textwidth}{!}{%
        \begin{tabular}{lcc}
        \hline
         & \textbf{Accuracy} & \textbf{Idn} \\
         \hline
         \textbf{HiTAB} \\
         \cline{1-1}
         Table {\textsubscript{\textit{w.} spt \& \textit{w/o.} ref}} & 20.3 & 9.9 \\
         Table {\textsubscript{\textit{w.} mpt \& \textit{w.} ref}} & 49.3 & 0.6\\
         \hline
         \textbf{AIT-QA}\\
         \cline{1-1}
         Table {\textsubscript{\textit{w.} spt \& \textit{w/o.} ref}} & 64.7 & 14.2 \\
         Table {\textsubscript{\textit{w.} mpt \& \textit{w.} ref}} & 76.3 & 0.2\\
         \hline
        \end{tabular}
    }
    \centering
    \caption{Ablation sresults on HiTAB and AIT-QA. \textit{spt} represents a simple prompt. \textit{mpt} represents the prompt that combines the information of the reconstructed table with the CoT design of Complex Table QA. \textit{ref} represents a JSON-formatted table. \textit{Idn} stands for "I don't know". We ask the model to answer "I don't know" if it could not infer the answer based on the existing context in the prompt. Please refer to Appendix \ref{Single-Turn details}, \ref{Multi-Turn details} and \ref{Simple prompt} for more details.}
    \label{tab:4}
\end{table}
}

\subsection{Ablation Study}
As shown in Table \ref{tab:4}, we conduct ablation experiments on the HiTAB and AIT-QA datasets. The results indicate that by restructuring the tables and integrating the restructured structural information with the CoT design prompts for the Complex Table QA task, the ability of GPT-3.5 to parse complex tables is significantly enhanced. Simultaneously, there is a notable reduction in the instances where the model responds with "I don't know." This demonstrates that our method can effectively improve the perception ability of the model regarding complex table structures, and consequently enhance its capacity to analyze complex tables.

\section{Analysis}
The results show that our method proposed for complex table parsing based on GPT-3.5 effectively outperforms the optimal methods on the existing two complex table datasets. However, compared to the significant performance with single-turn dialogue, the accuracy with multi-turn dialogue, especially on the HiTAB dataset, performs slightly worse. We further analyze the results in this section.

\subsection{Effects of the Input Token Limit}
Due to the inherent length of complex tables, as seen in Table \ref{tab:1}, filling them into prompt templates increases the likelihood of exceeding the input token limit of GPT-3.5, which has the following implications for our task.
\subsubsection{Context Truncation}
The excessive length of the input text leads to context truncation, resulting in the omission of critical information. As shown in Table \ref{tab:4}, the accuracy on the HiTAB dataset is quite low without table reconstruction, which is attributed to the fact that after filling the prompt template slot with the table information, a large number of prompts exceed the input token limit, and the direct truncation of the context leads to the missing of valuable information. As a solution, we employ a combination of single-turn and multi-turn dialogues to accomplish complex table QA.
\subsubsection{Trade-off between Depth and Breadth}
When opting for single-turn dialogue for Complex Table QA, we input all information at once.   The model is exposed to a wide and comprehensive range of information sources, but due to the absence of interactive prompts, its ability to focus on key information in the questions and tables is limited.
In contrast, when employing multi-turn dialogue, through continuous interactive prompting, we consistently guide the model to identify key information from each dialogue turn.  We then utilize this information to further prompt the model in subsequent turns. Multi-turn dialogue enables the model to be more focused on key information, but there is a possibility that it may overlook the global context.

Additionally, given that GPT-3.5 cannot automatically retain historical conversations, errors accumulate as dialogue turns increase, rendering the answers increasingly prone to inaccuracy.

\subsubsection{Prompt Design Limitation}
\cite{dialogue_understanding} pointed out that providing both examples and descriptions in prompt can effectively improve the performance of ChatGPT. However, due to the length of the complex table, we cannot provide an example with  complete dialogue and description in the prompt concurrently.
\begin{figure}[t]
    \centering
    
    \includegraphics[width=0.45\textwidth]{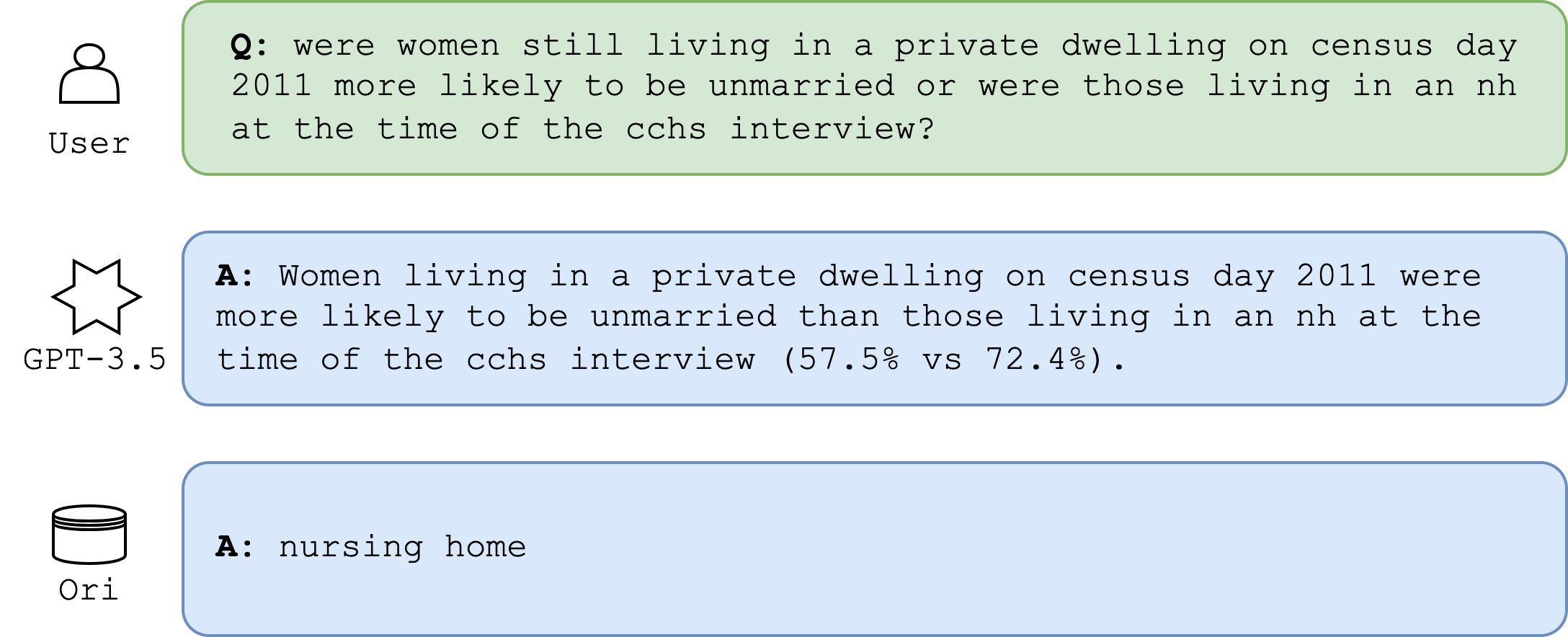}
    \caption{An example of hallucination of GPT-3.5.}
    \label{fig:hallu}
\end{figure}
\subsection{Effect of Hallucination}
Hallucination refers to the generation of content that may seem plausible but is not grounded in the source information or is outright incorrect. As seen in the Figure \ref{fig:hallu}, when we pose the question, \textit{"were women still living in a private dwelling on census day 2011 \textbf{more likely} to be unmarried or were those living in an nh at the time of the cchs interview?"}, GPT-3.5 responded with \textit{"Women"} and further provided a detailed explanation stating \textit{"(57.5\% vs 72.4\%)"}. Evidently, if we make our selection based on the analysis provided in the answer, the appropriate response would be \textit{"women living in an n"}, which refers to a \textit{"nursing home"}.

\subsection{Effect of Hierarchical Row Header}
In conjunction with Table \ref{tab:3}, it can be observed that the performance of not only our method but also others on the subset "Row header hierarchy" begins to decline when the answer relies on the row header hierarchy compared with the subset "No row header hierarchy".

During the analysis of experimental results on the HiTAB dataset, we also encountered the same issue: the model exhibits different levels of attention to row headers and column headers in tables. Under the same data format and prompt conditions, GPT-3.5 typically tends to focus more on column headers. We consider that this behavior might be attributed to the inherent ability of GPT-3.5 to interpret tables in markdown format, which typically only has column headers or places emphasis on them. Consequently, when attempting to understand tables in a new format, GPT-3.5 may transfer its prior knowledge, thereby affecting its parsing of the new tables.

\subsection{Error Analysis}
We randomly sample 50 instances, each from the erroneous results of single-turn dialogue and multi-turn dialogue, and analyze them in terms of the accuracy in the row and column header localization and cell selection. The results indicate that the errors are primarily attributed to the model either locating incorrect row and column headers or locating too few row and column headers.

In the analysis of single-turn dialogue and multi-turn dialogue, it can be found that  errors are mainly caused by incorrect localization of row and column headers. When the model locates non-header tuples based on incorrect or incomplete row and column header tuples, it tends to make erroneous selections. Furthermore, when the model selects a broad range of row and column headers, the number of non-header tuples that are selected based on them increases. The excess information can also lead to incorrect selections (another main reason for the erroneous answers of the model). This suggests that the comprehension capability of GPT-3.5 might decline when the input content is lengthy and complex. In addition, the model sometimes chooses the wrong cells even under the guidance of correct row and column headers, such as generating fictitious answers on their own, indicating that GPT-3.5 is sometimes prone to hallucinations.

\section{Conclusion}
Inspired by the powerful reasoning and generation capabilities of LLMs, we propose to leverage GPT-3.5 as a parser for complex tables with table reformat and prompt design. Extensive experiments show that our method achieves the best performance on HiTAB and AIT-QA. However, limited by the input length and inability to store historical information of GPT-3.5, how to utilize multi-turn dialogue to obtain more valuable context for Complex Table QA remains to be explored.

\section*{Limitations}
When utilizing GPT-3.5, a generative model, for the Complex Table QA task, comparing its generated outputs to the original answers is an important issue.
\begin{itemize}
    \item GPT-3.5 typically generates answers in its own words. When paraphrasing the information from the context or that it has learned during training, it may provide details that do not align in granularity with the original answer. 

    \item GPT-3.5 is unstable. When posing the same question within the same context, it does not always provide consistent responses.

    \item GPT-3.5 is prone to hallucinations. It is capable of generating answers that bear no relevance to the original text or the original answer.
\end{itemize}

Given these issues, it is important to define an evaluation metric that takes into account the complexity and variability of the responses generated by GPT-3.5.


\section*{Acknowledgements}
We deeply indebted to the anonymous reviewers from EMNLP for their constructive feedback. We are also express our gratitude to our tutors and peers for their invaluable insights and unflagging support throughout the course of this research. This work is also supported by National Natural Science Foundation of China (No.62172101); and by the Science and Technology Commission of Shanghai Municipality (No.22511106000); and Regional social experimentation with the "Clinical Decision Support System for Paediatric Outpatient Clinics" (PROJECT NO: 21002411800); and Auxiliary Diagnosis and Rare Disease Screening System for Children's Pneumonia (No.yg2022-7).
\bibliography{anthology,custom}
\bibliographystyle{acl_natbib}

\appendix

\section{Appendix}
\label{sec:appendix}

\subsection{Single-Turn Prompt}
\label{Single-Turn details}
The \textbf{Regulation} and \textbf{Input Context} parts of the single-turn dialogue prompt are detailed in Tables \ref{tab:sig_rgl} and \ref{tab:sig_ctt}, respectively.

\begin{table*}[h]
    \centering
    \begin{tabular}{m{15cm}}
    
    \hline
     \textbf{Regulations} \\
    \midrule
     \# \textit{Role play} \\
     Suppose you are an expert in statistical analysis.\\
     
     You will be given a table described in a special format.\\
     
     Your task is to answer the questions based on the content of the table. \\

     \hline
     
     \# \textit{Table Description} \\
     The table is described as follows:\\
     1. The title means the title of the table.\\

     2. A tuple (T, T1, T2, T3, T4) represents a column header, where T indicates it's a column header, T1 denotes its level, T2 and T3 indicate the start and end column of the header, and T4 specifies the content.\\

     3. A tuple (L, L1, L2, L3, L4) represents a row header, where L indicates it's a row header, L1 denotes its level, L2 and L3 indicate the start and end row of the header, and L4 specifies the content.\\

     4. We represent non-header tuples as (C, C1, C2, C3), where C denotes a non-header tuple, C1 denotes the row, C2 denotes the column, and C3 denotes the content. C1 corresponds to the row header tuple’s L2 and L3 that are related to it, while C2 corresponds to the column header tuple’s T2 and T3 that are related to it.\\

     5. The tuple of a column header contains T1, representing the level of the header, with 0 being the highest level and larger T1 indicating lower levels. If the T2 and T3 of tuple A are between T2 and T3 of tuple B (can be equal), then there is a parent-child relationship between A and B, A is a sub-header of B, B is a parent-header of A, and A's T1 must be smaller than B's T1. The lowest level header's tuple has T2=T3. Similarly for row headers. The specific tuples are in Table Content.\\

     \hline

     \# \textit{Examples} \\
     For examples:\\
    
     The tuple (T, 1, 0, 0, g) denotes a column header with level 1, spanning from column 0 to column 0,  with the content "g".\\
    
     The tuple (L, 0, 6, 6, karlsruher sc) denotes a row header with level 0, spanning from row 6 to row 6, with the content "karlsruher sc".\\
  
     The tuple (C, 7, 0, 416) represents a non-header cell at row 7, column 0, with a value of 416.\\
    
     Make sure you read and understand these instructions carefully. \\

     \hline

     \# \textit{Chain-of-thought}

     Let's think step by step as follows and give full play to your expertise as a statistical analyst: \\

     1. Clearly understand the question and the information needed to answer the question to determine the necessary information to extract. \\

     2. Have a comprehensive understanding of the data in the table, including the meaning, data types, and formats of each column and row tuples (Note: There are usually summative tuples in the table, such as all, combine, total, sum, average, mean, etc. These tuples help you skip a lot of operations).\\
    
     3. Based on the question, select the row and column header tuples that are most relevant to the question and then locate the non-header tuples based on the row and column header tuples you selected before.\\
    
     4. Perform statistical, calculation, sorting, grouping, or other operations on the tuples you selected before to extract useful information based on the question's requirements.\\
     \hline
     \# \textit{Output Control}\\
    
     You MUST answer each question in the format below line by line (Note: Keep your answer concise):\\
    
     1. Column header: The column header tuples most relevant to the answer.\\
    
     2. Row header: The row header tuples most relevant to the answer.\\
    
     3. Cell: The non-header tuples most relevant to the answer.\\
    
     4. Operation: the operation you performed on the tuples you selected.\\
    
     5. Answer: your answer (A number, noun, phrase, or set of data).\\
    
     And if the answer is not contained within the context, say "I don't know".\\
     \hline
    \end{tabular}
    \caption{Full text of the "Regulations" part of the single-turn dialogue prompt. The text in "[]" can be replaced according to the specific information in the QA process.}
    \label{tab:sig_rgl}
\end{table*}

\begin{table*}[t]
    \centering
    \begin{tabular}{m{15cm}}
        \hline
        \textbf{Input Context}\\
        \hline
        \# \textit{Table Title}\\
        Title: [TABLE\_TITLE\_HERE]\\
        \# \textit{Column Header}\\
        Column header: [TABLE\_COLUMN\_HEADER\_HERE]\\
        \# \textit{Row Header}\\
        Row header: [TABLE\_ROW\_HEADER\_HERE]\\
        \# \textit{Non-Header}\\
        Non-header: [TABLE\_NON\_HEADER\_HERE]\\
        \# \textit{Question}\\
        Q: [QUSTION\_HERE]\\
        A:
        \\
        \hline
        
    \end{tabular}
    \caption{Full text of the "Input Context" part of the single-turn dialogue prompt. The text in "[]" can be replaced according to the specific information in the QA process.}
    \label{tab:sig_ctt}
\end{table*}

\subsection{Multi-Turn Prompt}
\label{Multi-Turn details}

As shown in \ref{tab:turn1}, \ref{tab:turn2} and \ref{tab:turn3}, we describe the prompt of the first, second and third prompt turn in detail.
\begin{table*}[h]
    \centering
    \begin{tabular}{m{15cm}}
    
    \hline
     \textbf{Regulations} \\
    \midrule
     \# \textit{Role play} \\
     Suppose you are an expert in statistical analysis.\\
     
     You will be given a table described in a special format.\\
     
     Your task is to answer the questions based on the content of the table. \\

     \hline
     
     \# \textit{Table Description} \\
     The table is described as follows:\\
     1. The title means the title of the table.\\

     2. A tuple (T, T1, T2, T3, T4) represents a column header, where T indicates it's a column header, T1 denotes its level, T2 and T3 indicate the start and end column of the header, andT4 specifies the content.\\

     3. A tuple (L, L1, L2, L3, L4) represents a row header, where L indicates it's a row header, L1 denotes its level, L2 and L3 indicate the start and end row of the header, and L4 specifies the content.\\

     4. We represent non-header tuples as (C, C1, C2, C3), where C denotes a non-header tuple, C1 denotes the row, C2 denotes the column, and C3 denotes the content. C1 corresponds to the row header tuple’s L2 and L3 that are related to it, while C2 corresponds to the column header tuple’s T2 and T3 that are related to it.\\

     5. The tuple of a column header contains T1, representing the level of the header, with 0 being the highest level and larger T1 indicating lower levels. If the T2 and T3 of tuple A are between T2 and T3 of tuple B (can be equal), then there is a parent-child relationship between A and B, A is a sub-header of B, B is a parent-header of A, and A's T1 must be smaller than B's T1. The lowest level header's tuple has T2=T3. Similarly for row headers. The specific tuples are in Table Content.\\

     \hline

     \# \textit{Examples} \\
     For examples:\\
    
     The tuple (T, 1, 0, 0, g) denotes a column header with level 1, spanning from column 0 to column 0,  with the content "g".\\
    
     The tuple (L, 0, 6, 6, karlsruher sc) denotes a row header with level 0, spanning from row 6 to row 6, with the content "karlsruher sc".\\
  
     The tuple (C, 7, 0, 416) represents a non-header cell at row 7, column 0, with a value of 416.\\
    
     Make sure you read and understand these instructions carefully. \\

     \hline

     \textbf{Input Context \#1}\\
     \hline
     \# \textit{Abstract Keywords}\\
     Extract the key words in the question.\\
    
     Q: [QUESTION\_HERE]\\

     A:  \\
    \hline
     
    \end{tabular}
    \caption{Full text of the first prompt turn of the multi-turn dialogue. The text in "[]" can be replaced according to the specific information in the QA process.}
    \label{tab:turn1}
\end{table*}

\begin{table*}[h]
    \centering
    \begin{tabular}{m{15cm}}
    
    \hline
     \textbf{Regulations and Historical Dialogue} \\
    \midrule
     \# \textit{Role play} \\
     Suppose you are an expert in statistical analysis.\\
     
     You will be given a table described in a special format.\\
     
     Your task is to answer the questions based on the content of the table. \\

     \hline
     
     \# \textit{Table Description} \\
     The table is described as follows:\\
     1. The title means the title of the table.\\

     2. A tuple (T, T1, T2, T3, T4) represents a column header, where T indicates it's a column header, T1 denotes its level, T2 and T3 indicate the start and end column of the header, andT4 specifies the content.\\

     3. A tuple (L, L1, L2, L3, L4) represents a row header, where L indicates it's a row header, L1 denotes its level, L2 and L3 indicate the start and end row of the header, and L4 specifies the content.\\

     4. We represent non-header tuples as (C, C1, C2, C3), where C denotes a non-header tuple, C1 denotes the row, C2 denotes the column, and C3 denotes the content. C1 corresponds to the row header tuple’s L2 and L3 that are related to it, while C2 corresponds to the column header tuple’s T2 and T3 that are related to it.\\

     5. The tuple of a column header contains T1, representing the level of the header, with 0 being the highest level and larger T1 indicating lower levels. If the T2 and T3 of tuple A are between T2 and T3 of tuple B (can be equal), then there is a parent-child relationship between A and B, A is a sub-header of B, B is a parent-header of A, and A's T1 must be smaller than B's T1. The lowest level header's tuple has T2=T3. Similarly for row headers. The specific tuples are in Table Content.\\

     \hline

     \# \textit{Examples} \\
     For examples:\\
    
     The tuple (T, 1, 0, 0, g) denotes a column header with level 1, spanning from column 0 to column 0,  with the content "g".\\
    
     The tuple (L, 0, 6, 6, karlsruher sc) denotes a row header with level 0, spanning from row 6 to row 6, with the content "karlsruher sc".\\
  
     The tuple (C, 7, 0, 416) represents a non-header cell at row 7, column 0, with a value of 416.\\
    
     Make sure you read and understand these instructions carefully. \\

     \hline

     \# \textit{Output of Turn 1 }\\
     Extract the key words in the question.\\
    
     Q: [QUESTION\_HERE]\\

     A: [ANSWER\_OF\_TURN\_1] \\
    \hline

    \textbf{Input Context \#2}\\
    \hline

    \# \textit{Select Headers}\\
    Here are table title and the tuples of the table's rows and headers, please try to locate the lowest level of headers that match the question and keywords you extracted:\\

    "Title": [TABLE\_TITLE\_HERE]\\

    "Column header": [TABLE\_COLUMN\_HEADER\_HERE]\\

    "Row header": [TABLE\_ROW\_HEADER\_HERE]\\

    \# \textit{Output Control} \\
    You MUST output your selection in the following format:\\

    1. Column header: \\

    2. Row header:\\

\hline
     
    \end{tabular}
    \caption{Full text of the second prompt turn of the multi-turn dialogue. The text in "[]" can be replaced according to the specific information in the QA process.}
    \label{tab:turn2}
\end{table*}

\begin{table*}[h]
    \centering
    \begin{tabular}{m{15cm}}
    
    \hline
     \textbf{Regulations and Historical Dialogue} \\
    \midrule
     \# \textit{Role play} \\
     Suppose you are an expert in statistical analysis.\\
     
     You will be given a table described in a special format.\\
     
     Your task is to answer the questions based on the content of the table. \\

     \hline
     
     \# \textit{Table Description} \\
     The table is described as follows:\\
     1. The title means the title of the table.\\

     2. A tuple (T, T1, T2, T3, T4) represents a column header, where T indicates it's a column header, T1 denotes its level, T2 and T3 indicate the start and end column of the header, andT4 specifies the content.\\

     3. A tuple (L, L1, L2, L3, L4) represents a row header, where L indicates it's a row header, L1 denotes its level, L2 and L3 indicate the start and end row of the header, and L4 specifies the content.\\

     4. We represent non-header tuples as (C, C1, C2, C3), where C denotes a non-header tuple, C1 denotes the row, C2 denotes the column, and C3 denotes the content. C1 corresponds to the row header tuple’s L2 and L3 that are related to it, while C2 corresponds to the column header tuple’s T2 and T3 that are related to it.\\

     5. The tuple of a column header contains T1, representing the level of the header, with 0 being the highest level and larger T1 indicating lower levels. If the T2 and T3 of tuple A are between T2 and T3 of tuple B (can be equal), then there is a parent-child relationship between A and B, A is a sub-header of B, B is a parent-header of A, and A's T1 must be smaller than B's T1. The lowest level header's tuple has T2=T3. Similarly for row headers. The specific tuples are in Table Content.\\

     \hline

     \# \textit{Examples} \\
     For examples:\\
    
     The tuple (T, 1, 0, 0, g) denotes a column header with level 1, spanning from column 0 to column 0,  with the content "g".\\
    
     The tuple (L, 0, 6, 6, karlsruher sc) denotes a row header with level 0, spanning from row 6 to row 6, with the content "karlsruher sc".\\
  
     The tuple (C, 7, 0, 416) represents a non-header cell at row 7, column 0, with a value of 416.\\
    
     Make sure you read and understand these instructions carefully. \\

     \hline

     \# \textit{Output of Turn 2 and Code}\\
     Here are all the tuples relevant to the question:\\

     [ANSWER\_OF\_TURN\_2] \\
     Non-header: [OUTPUT\_OF\_CODE]\\

    \hline

    \textbf{Input Context \#2}\\

    \# \textit{All Headers}\\

    "Column header": [TABLE\_COLUMN\_HEADER\_HERE]\\

    "Row header": [TABLE\_ROW\_HEADER\_HERE]\\

    \# \textit{Output Control}\\
    
     You MUST answer each question in the format below line by line (Note: Keep your answer concise):\\
    
     1. Column header: The column header tuples most relevant to the answer.\\
    
     2. Row header: The row header tuples most relevant to the answer.\\
    
     3. Cell: The non-header tuples most relevant to the answer.\\
    
     4. Operation: the operation you performed on the tuples you selected.\\
    
     5. Answer: your answer (A number, noun, phrase, or set of data).\\
    
     And if the answer is not contained within the context, say "I don't know".\\

     Notes:\\

     1. In the row and column header tuples, the third and fourth elements represent the row and column position information.\\
     
     2. You must output non-header tuples that are valid in the above tuples.\\

     \hline

    \end{tabular}
    \caption{Full text of the third prompt turn of the multi-turn dialogue. The text in "[]" can be replaced according to the specific information in the QA process.}
    \label{tab:turn3}
\end{table*}

\subsection{Simple Prompt}
\label{Simple prompt}
\begin{flushleft}
 Table \ref{tab:simple} shows the full text of a specific prompt designed for the original table format.
\begin{table*}[t]
    \centering
    \begin{tabular}{m{15cm}}
        \hline
        \textbf{Regulations}\\
        \hline
        The text provided describes a table in json format, answer the question as truthfully as possible using the provided text and don't omit the decimal point, and if the answer is not contained within the text below, say "I don't know".\\
        \hline
        \# Content
        [ORIGINAL\_TABLE]\\
        Q: [QUESTION\_HERE]\\
        A: \\
        \hline
    \end{tabular}
    \caption{Full text of the simple prompt. The text in "[]" can be replaced according to the specific information in the QA process.}
    \label{tab:simple}
\end{table*}

\end{flushleft}

\end{document}